\documentclass[letterpaper]{article} 
\usepackage[]{aaai23}  
\usepackage{times}  
\usepackage{helvet}  
\usepackage{courier}  
\usepackage[hyphens]{url}  
\usepackage{graphicx} 
\usepackage{natbib}  
\usepackage{caption} 
\usepackage{algorithm}
\usepackage{algorithmic}

\usepackage{newfloat}
\usepackage{listings}

\usepackage{booktabs}
\usepackage{multirow}
\usepackage{caption}
\usepackage{cite}
\usepackage{pifont}
\usepackage{hyperref}
\usepackage{appendix}

\DeclareCaptionStyle{ruled}{labelfont=normalfont,labelsep=colon,strut=off} 
\floatstyle{ruled}
\newfloat{listing}{tb}{lst}{}
\floatname{listing}{Listing}
\title{Boosting Point Clouds Rendering via Radiance Mapping}

\author{
    Xiaoyang Huang\textsuperscript{\rm 1}\equalcontrib, 
    Yi Zhang\textsuperscript{\rm 1}\equalcontrib, 
    Bingbing Ni\textsuperscript{\rm 1}\thanks{Corresponding Author.}, 
    Teng Li\textsuperscript{\rm 2}, 
    Kai Chen\textsuperscript{\rm 3},
    Wenjun Zhang\textsuperscript{\rm 1}
}
\affiliations{
    \textsuperscript{\rm 1}Shanghai Jiao Tong University, Shanghai 200240, China, \\
    \textsuperscript{\rm 2}Anhui University, 
    \textsuperscript{\rm 3}Shanghai AI Lab

    \{huangxiaoyang, yizhangphd, nibingbing\}@sjtu.edu.cn
}

\usepackage{bibentry}

\begin{document}

\maketitle

\begin{abstract}
Recent years we have witnessed rapid development in NeRF-based image rendering due to its high quality. However, point clouds rendering is somehow less explored. Compared to NeRF-based rendering which suffers from dense spatial sampling, point clouds rendering is naturally less computation intensive, which enables its deployment in mobile computing device. In this work, we focus on boosting the image quality of point clouds rendering with a compact model design. We first analyze the adaption of the volume rendering formulation on point clouds. Based on the analysis, we simplify the NeRF representation to a spatial mapping function which only requires single evaluation per pixel. Further, motivated by ray marching, we rectify the the noisy raw point clouds to the estimated intersection between rays and surfaces as queried coordinates, which could avoid \textit{spatial frequency collapse} and neighbor point disturbance. Composed of rasterization, spatial mapping and the refinement stages, our method achieves the state-of-the-art performance on point clouds rendering, outperforming prior works by notable margins, with a smaller model size. We obtain a PSNR of 31.74 on NeRF-Synthetic, 25.88 on ScanNet and 30.81 on DTU. Code and data are publicly available\footnote{\url{https://github.com/seanywang0408/RadianceMapping}}.
\end{abstract}

\begin{figure*}[h!]
\begin{center}
  \includegraphics[width=\linewidth]{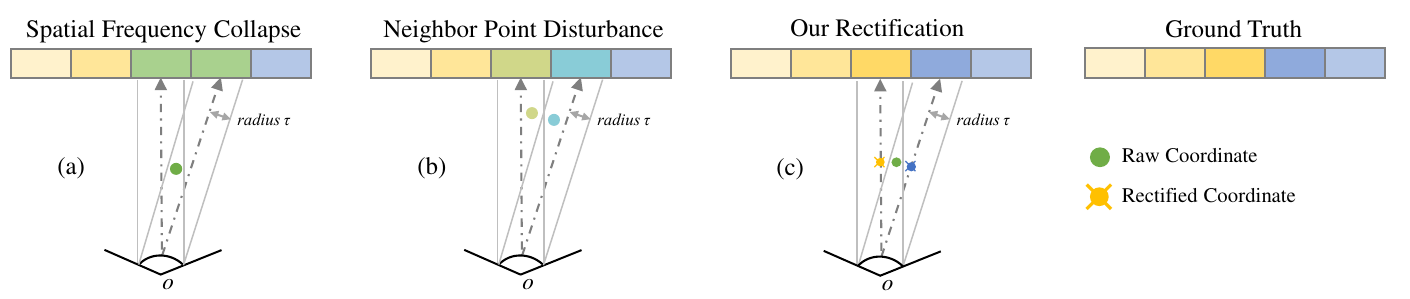}
\end{center}
\vspace{-8pt}
\caption{\textbf{(a)} \textit{Spatial frequency collapse} occurs when using neural descriptors or raw point clouds query. The point is optimized to a green color, which is the mixing of yellow and blue. \textbf{(b)} Using raw point clouds query would additionally cause neighbor point disturbance. The points lying close have a larger impact on the feature optimization of each other. \textbf{(c)} Our coordinate rectification could alleviate the above issues. The idea is illustrated in 1D rendering.}
\label{fig:collapse}
\end{figure*}

\section{Introduction}

The rising trend of AR/VR application calls for better image quality and higher computation efficiency in rendering technology. Recent works mainly focus on NeRF-based (\citeauthor{mildenhall2020nerf}) rendering due to its photo-realistic effect. Nevertheless, NeRF-based rendering suffers from heavy computation cost, since its representation assumes no explicit geometry is known, and requires burdensome spatial sampling. This drawback severely hampers its application in mobile computing devices, such as smart phones or AR headsets. On the other hand, point clouds (\citeauthor{huang2022representation}), which have explicit geometry, are easy to obtained as the depth sensors become prevalent and MVS algorithms (\citeauthor{yao2018mvsnet, wang2021patchmatchnet}) get powerful. It deserves more attention to develop high-performance rendering methods based on point clouds, which is so far insufficiently explored. In this work, we introduce a point clouds rendering method which achieves comparable rendering performance to NeRF.


The main difference between NeRF-based rendering and point clouds rendering is that the latter is designed upon the noisy surface of objects. On the bright side, it is a beneficial geometric prior which could greatly reduce the query times in 3D space. On the bad side, this prior is noisy and sparse, since the point clouds are generally reconstructed by MVS algorithms or collected by depth sensors. It needs additional approaches to alleviate the artifact brought by the noise and sparsity. Therefore, most of the current point clouds rendering methods require two steps. One is the spatial feature mapping, and the other is image-level refinement. The spatial feature mapping step is similar to the NeRF representation, which maps a 3D coordinate to its color, density or latent feature. The refinement step is usually implemented as a convolutional neural network. In this work, we mainly focus on the spatial feature mapping step. Previous works use point clouds voxelization (\citeauthor{dai2020neural}), learnable parameters (\citeauthor{ruckert2021adop, kopanas2021point}) or linear combination of sphere basis (\citeauthor{rakhimov2022npbg++}) as mapping functions. However, these methods suffer either from high computation cost, large storage requirements, or unsatisfactory rendering performance. To this end, we introduce a much simpler but surprisingly effective mapping function. Motivated by the volume rendering formulation in NeRF, we analyze its adaptation on point clouds rendering scenarios. It is concluded that in a point cloud scene, the volumetric rendering could be simplified to the modeling of the view-dependent color of the first-time intersection between the estimated surface and the ray. In other words, we augment each 3D point (i.e., most probably a surface point) with a learnable feature indicating first-hit color. Thereby the point clouds rendering task could be re-cast within the high fidelity NeRF framework, without consuming redundant computation on internal ray samples. We name it \textit{radiance mapping}. Moreover, based on radiance mapping, we rectify the raw point cloud coordinates that are fed into the mapping function using the z-buffer in rasterization to obtain a query point which lies exactly on the camera ray. This approach allows us to obtain a more accurate geometry and avoid spatial frequency collapse. The radiance mapping function consisted of a 5-layer MLP is only 0.75M large, which is much smaller than the spatial feature mapping functions in previous works, but with notably better performance. Followed by a 2D neural renderer which compensates the sparcity and noise in point clouds as done in previous works, our complete model is approximately 8M in total. 

Our method reaches comparable rendering effect compared to NeRF, but with much less computation cost since it only needs single model inference per pixel. Compared to prior point clouds rendering methods, we obtain notable improvement in terms of image quality, with a smaller model size and simpler computation. We achieve a PSNR of 31.74 on NeRF-Synthetic (\citeauthor{mildenhall2020nerf}), 25.88 on ScanNet (\citeauthor{dai2017scannet}) and 30.81 on DTU (\citeauthor{aanaes2016large}). As far as we know,  It is the state-of-the-art result on this task.

\section{Related Work}

\subsection{Implicit Rendering}
\paragraph{NeRF-based} Neural Radiance Fields (\citeauthor{mildenhall2020nerf}) advance the neural rendering quality to a higher level. NeRF represents the scene using an MLP which predicts the color and density of a point. It projects the points along the camera ray to the pixel color with volume rendering. Following NeRF, there are various innovations which address the different challenges in NeRF representation. PixelNeRF (\citeauthor{yu2021pixelnerf}), IBRNet (\citeauthor{wang2021ibrnet}) and DietNeRF (\citeauthor{jain2021putting}) render novel views from only one or a few input images. NeRF-W (\citeauthor{martin2021nerf}) tackles the variable illumination and transient occluders in the wild instead of a static scene. Mip-NeRF (\citeauthor{barron2021mip}) and Mip-NeRF 360 (\citeauthor{barron2021mip}) improves the image quality by rendering anti-aliased conical frustums. NSVF (\citeauthor{liu2020neural}), PlenOctrees (\citeauthor{yu2021plenoctrees}) and TensoRF (\citeauthor{chen2022tensorf}) aim at accelerating the inference speed of NeRF by building a more efficient structure after scene fitting. Point-NeRF (\citeauthor{xu2022point}) also assumes a point cloud is given like ours. But it still follows the volume rendering formulation in NeRF, which also suffers from dense spatial sampling.
\paragraph{Implicit Surface Rendering} This line of works aim at reconstruction the implicit surfaces via neural rendering. DVR (\citeauthor{niemeyer2020differentiable}) learns implicit 3D representation from images by analytically deriving depth gradients from implicit differentiation. IDR (\citeauthor{yariv2020multiview}) renders an implicit surface by appromitaing the light reflected from the surface towards the camera. UNISURF (\citeauthor{oechsle2021unisurf}) combines implicit surface models and radiance fields together to enable surface reconstruction without object masks. NeuS (\citeauthor{wang2021neus}) gives a theoretical proof that the classic volume rendering formulation causes error on the expectation of the object surface, and presents a solution which yields an unbiased SDF representation. \citeauthor{yariv2021volume} models the volume density as the function of the SDF representation, leading to a more accurate sampling of the camera ray.

\begin{figure*}[h!]
    \centering
    \includegraphics[width=\linewidth]{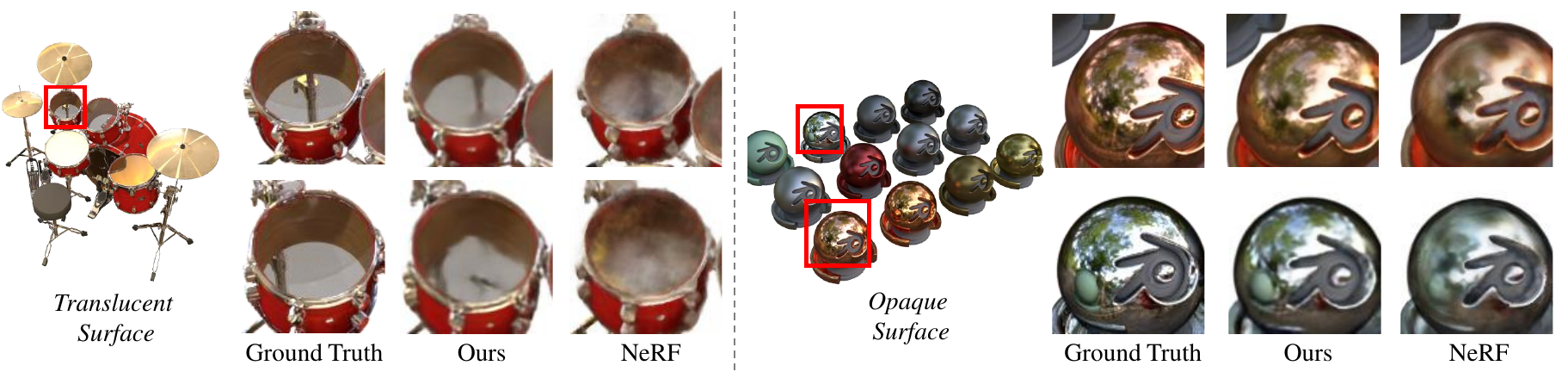}
    \captionof{figure}{The translucent and opaque surfaces rendered by NeRF and our method. In the Drums scene, both methods are optimizing the color of the first-intersected surface instead of modeling the correct translucency. The second row shows the membrane from another view. In the Materials scene, our method could even render more decorate specular effect on the smooth metal balls, while NeRF generates somehow blurry artifacts. The visualization in the original NeRF paper (\citeauthor{mildenhall2020nerf}) reveals the same artifact. We owe this superiority to the \textbf{explicit geometry} provided by point clouds.}
    \label{fig:drum}
\end{figure*}
\subsection{Point Clouds Rendering}

\paragraph{Inverse Rendering} Early work (\citeauthor{zwicker2001surface}) proposes a point cloud rendering method using an Elliptical Weighted Average filter based on Gaussian Kernel. \citeauthor{yifan2019differentiable} enables backward propagation of surface splatting to optimize the position of point clouds to match to object geometry from images. \citeauthor{insafutdinov2018unsupervised} use a differentiable point clouds projection module to unsupervisedly learn the object shape and pose from two-view images. \citeauthor{lin2018learning} propose pseudo-rendering which upsamples the target image to alleviate the collision effect in discretization. \citeauthor{wiles2020synsin} construct a point cloud from single-view images by using a depth regressor and spatial feature predictor, and render the point cloud with $\alpha$-composition followed by a ResNet (\citeauthor{he2016deep}) refinement network. The training is supervised by a photometric loss and a GAN loss (\citeauthor{wang2018high}). \citeauthor{zhou2017unsupervised} and \citeauthor{godard2017unsupervised} adopt a similar approach, but on a monocular depth estimation task with street-view video sequences. 
\paragraph{View Synthesis} 

NPBG (\citeauthor{aliev2020neural}) proposes to render novel views of a scene using point-based learnable neural descriptors and a U-Net refinement network. It adopts multi-scale rasterization to model image details in different level. \citeauthor{johnson2016perceptual, dosovitskiy2016generating} use a perceptual loss to optimize the network. \citeauthor{dai2020neural} propose to project the point clonds to a layered volume by voxelization. Then a 3D CNN (\citeauthor{maturana2015voxnet, yang2021reinventing}) is used to transform the volume into a set of multi-plane images and their blending weights, which form the final image by its the weighted sum. NPBG++ (\citeauthor{rakhimov2022npbg++}) reduces the running time upon NPBG, using a feature extracter to lift the neural descriptor feature and making it view-dependent. ADOP (\citeauthor{ruckert2021adop}) renders HDR and LDR images with a coarsely-initialized point cloud and camera parameters. The point clouds, camera poses and the 2D refinement network are jointly optimized. \citeauthor{kopanas2021point} perform scene optimization for each view based on bi-directional Elliptical Weighted Average splatting. \citeauthor{DBLP:journals/corr/abs-2112-01473} promote point clouds to implicit light fields to allow fast inference in view synthesis. READ (\citeauthor{li2022read}) adopt a similar approach to NPBG++ to synthesize photo-realistic street views for autonomous driving. We analyze the most relevant works to ours in the next section.







\section{Method}




\subsection{Spatial Mapping}
We first analyze the spatial mapping functions in previous point clouds rendering methods. Then we introduce our radiance mapping, a simpler but more effective mapping.

\paragraph{Previous Mapping Functions Revisited} NPBG (\citeauthor{aliev2020neural}) attachs learnable parameters to each point as neural descriptors. The advantage of this approach is that each point feature is optimized independently, and would not be influenced by nearby point feature. This is beneficial to those scenes where surface color changes drastically. However, it also leads to a drawback that the density of point clouds impose restrictions on the representation capacity of point feature. When the point clouds are sparse, the same neural descriptor would be rasterized to multiple pixels and optimized to fit the average of multiple pixels, which harms the rendering quality. We illustrate this issue in 1D rendering in Figure \ref{fig:collapse} (a). The point is optimized to a green color, which is the mixing of yellow and blue. Since the cause of the phenomenon is analogous to the dissatisfaction of Nyquist Rate in signal processing, we call it \textit{spatial frequency collapse}. On the other hand, when point clouds get comparatively dense due to higher quality of reconstruction or  depth sensor, the size of point feature would grows proportionally, which consumes more memory for storage and training. Besides, some of the point features which are only visible in a few views might not be sufficiently optimized.

\citeauthor{dai2020neural} propose to use a 3D CNN to extract spatial feature. It first voxelizes the point clouds into a layered volume, and then adopts a 3D CNN to extract spacial feature. Due to the high computation complexity of 3D CNN, this model is much more heavier, and not easy to deploy.

NPBG++ (\citeauthor{rakhimov2022npbg++}) develops a spatial mapping function motivated by sphere harmonics basis. It first uses a shared 2D CNN to extract image feature from multi-view images, and then aggregate the feature of each view into the point clouds by a linear combination of learnable basis functions over the unit sphere. This approach considers view direction as input, which would potentially generate better rendered images. However, it still suffers from proportionally increasing memory as the point clouds get more dense, similar to NPBG . Besides, it requires an additional U-Net as a image feature extractor which further increase model size.

\paragraph{Radiance Mapping} Comparing to the above spatial mapping functions, our method is much more light-weight. Our compact representation store the view-dependent radiance of the object surface. The idea is motivated from the volumetric rendering formulation in NeRF representations (\citeauthor{mildenhall2020nerf}), which take the 3D coordinate $\mathbf{x}=(x,y,z)$ and view direction $\mathbf{d}=(\theta, \phi)$ as inputs and output the color $\mathbf{c}$ and density $\sigma$ using a multi-layer perceptron (MLP) $\mathbf{F}_\Theta$, parameterized by $\Theta$:
\begin{equation}
    \mathbf{c}, \sigma = \mathbf{F}_\Theta(\mathbf{x}, \mathbf{d})
    \label{equ:nerf}
\end{equation}
Since NeRF representations assume no explicit geometry exists, each point lying on the camera ray $\mathbf{r}=\mathbf{o}+t\mathbf{d}$ are queried and aggregated to obtain the final pixel color $C(\mathbf{r})$:

\begin{equation}
    C(\mathbf{r})=\int_{t_{n}}^{t_{f}} T(t) \sigma(\mathbf{r}(t)) \mathbf{c}(\mathbf{r}(t), \mathbf{d}) dt
    \label{equ:volumetric_render}
\end{equation}
\begin{equation}
    T(t)=\exp \left(-\int_{t_{n}}^{t} \sigma(\mathbf{r}(s)) ds\right)
    \label{equ:transmittance}
\end{equation}


\begin{figure*}[!ht]
\begin{center}
  \includegraphics[width=\linewidth]{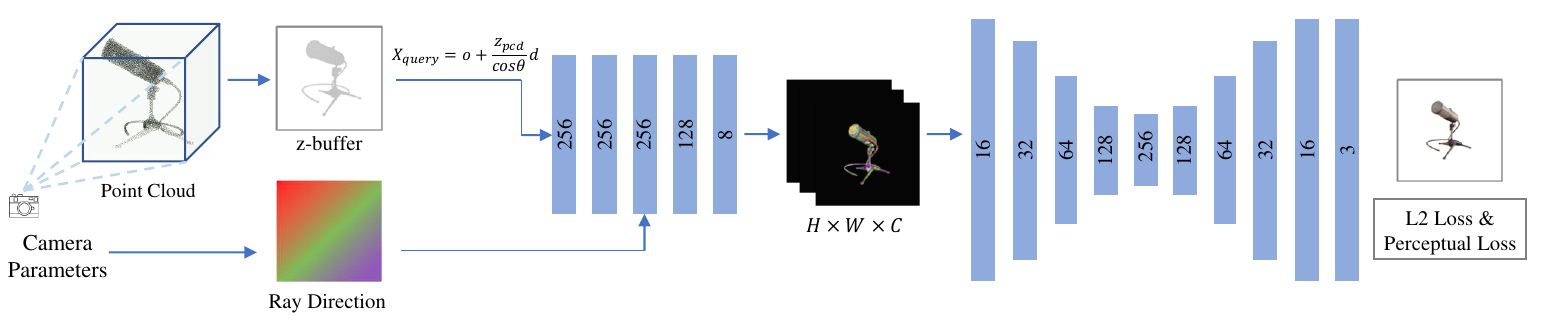}
\end{center}
\caption{Our end-to-end point clouds rendering pipeline. We use the z-buffer obtained from the rasterization stage to perform coordinate rectification. The rectified coordinates are fed into the radiance mapping MLP. The output feature map is sent to the refinement stage, which is implemented as a U-Net. We train the MLP and U-Net with a weighted sum of L2-loss and perceptual loss.}
\label{fig:method}
\end{figure*}

$T(t)$ denotes the accumulated transmittance along the ray. Suppose the camera ray first intersects with an object surface at point $\mathbf{p}_{o}=\mathbf{o}+t_o\mathbf{d}$. For those points that lie in front of $\mathbf{p}_{o}$, which satisfy $\{\mathbf{p}=\mathbf{o}+t\mathbf{d}|t<t_o\}$, would correspond to $\sigma_{o}=0$. Substituting it into Eqn. \ref{equ:transmittance} yields $\{T(t)=0|t<t_o\}$. Then we consider the points that lie behind $\mathbf{p}_{o}$, which satisfy $\{\mathbf{p}=\mathbf{o}+t\mathbf{d}|t>t_o\}$. One case is that the density of $\mathbf{p}_{o}$ equals to a comparatively large value, meaning that the object surface is opaque. In this case we have $\{T(t)\approx0|t>t_o\}$ according to Eqn. \ref{equ:transmittance}. Another case is that the density of $\mathbf{p}_{o}$ equals to a small value, in which the object surface is translucent. We demonstrate that the latter case could be discarded in point clouds rendering. \textit{(i)} No matter the point clouds are reconstructed by MVS algorithms or collected by depth sensors, only the first intersected surface would be revealed in the point clouds. It is impractical and not necessary to evaluate the surfaces behind the first one. \textit{(ii)} Even if only the first-intersected surface is evaluated, the correct pixel color could still be modeled in practice, since the point color $\mathbf{c}(\mathbf{r}(t_o), \mathbf{d})$ would be fitted to the pixel color $C(\mathbf{r})$ during optimization. Figure \ref{fig:drum} (left) shows the Drums scene in NeRF-Synthetic dataset, where the drum membrane is a translucent surface. We could see that the drum shelf behind the membrane is a thin column. If the translucency is modeled correctly, we should at least see the outline of the shelf. But in NeRF results we could see no outline but blurry artifact. It means that the density of the membrane predicted by NeRF is comparatively large such that the shelf behind is invisible. With no regard to the points that lie behind the intersection $p_o$, our method could still obtain similar results, or even better. Please refer to experiment section for more analysis. Given that $\{T(t)=0|t<t_o\}$ and $\{T(t)\approx0|t>t_o\}$ , the pixel color $C(\mathbf{r})$ becomes a single evaluation on $\mathbf{p}_{o}$ according to Eqn. \ref{equ:volumetric_render}. Thus we simply remove the density prediction in NeRF representation (Eqn. \ref{equ:nerf}). Moreover, we output the latent feature $\mathbf{L}$ that encodes the color to a higher dimension instead of RGB as prior works do. Then Eqn. \ref{equ:nerf} is reformulated as
\begin{equation}
    \mathbf{L} = \mathbf{F}_\Theta(\mathbf{x}, \mathbf{d})
    \label{equ:radiace_mapping}
\end{equation}
Since we do not need to predict the density, we reduce the number of MLP layers from $13$ in NeRF to $5$, yielding a model size as small as $0.75M$. This compact architecture enables fast inference and small storage requirement. As compared in Table \ref{tab:size_time}, our spatial mapping model is notably smaller than its counterparts in prior works. Since our model is light-weight with respect to model size and computation and needs only single time inference for each pixel, it is possible to deploy it in mobile computing device like smart phones or AR/VR headsets.

\begin{table*}[!ht]
    \centering
    \setlength\tabcolsep{1pt}
    \caption{Results on three datasets. For NeRF-Synthetic, the results are averaged on Ficus, HotDog and Mic, following NPBG++.}
    \vspace{-8pt}
    \begin{tabular*}{\hsize}{@{}@{\extracolsep{\fill}}lccccccccc@{}}
    \toprule
     & \multicolumn{3}{c}{NeRF-Synthetic} & \multicolumn{3}{c}{ScanNet }  & \multicolumn{3}{c}{DTU}  \\
     Method & 
     PSNR$\uparrow$ & SSIM$\uparrow$ & LPIPS$\downarrow$ & PSNR$\uparrow$ & SSIM$\uparrow$ & LPIPS$\downarrow$ & PSNR$\uparrow$ & SSIM$\uparrow$ & LPIPS$\downarrow$ \\
    \midrule
    NPBG &
    30.47 & 0.957 & 0.046 &
    25.08 & 0.737 & 0.459 &
    30.24 & 0.895 & 0.130 \\
    NPBG++&
    30.52 & 0.960 & 0.042 &
    25.24 & 0.755 & 0.444 & 
    30.25 & 0.894 & 0.131\\
    Ours &
    \textbf{31.74} & \textbf{0.968} & \textbf{0.029} &
    \textbf{25.88} & \textbf{0.794} & \textbf{0.414} &
    \textbf{30.81} & \textbf{0.904} & \textbf{0.128} \\
    \midrule
    NeRF & 
    33.07 & 0.989 & 0.064 &
    25.63 & 0.765 & 0.436 &
    31.03 & 0.909 & 0.127\\
    \bottomrule

    \end{tabular*}
    \label{tab:benchmark}
\end{table*}
    
\begin{table*}[!ht]
    \centering
    \setlength\tabcolsep{1pt}
    \caption{Per-scene PSNR in NeRF-Synthetic.}
    \vspace{-8pt}
    \begin{tabular*}{\hsize}{@{}@{\extracolsep{\fill}}lccccccccc@{}}
    \toprule
    Method 
    & Chair & Drums & Ficus & Hotdog 
    & Lego & Materials & Mic & Ship & Mean\\
    \midrule
  
    NPBG 
    & 28.81 & 23.57 & 28.23 & 32.03 & 27.72 & 27.24 & 31.16 & 26.04 & 28.10\\
    NPBG++ 
    & 28.72 & 23.60 & 28.11 & 32.22 & \textbf{27.84} & 27.12 & 31.23 & 26.11 & 28.12\\
    Ours 
    & \textbf{31.13} & \textbf{24.51} & \textbf{29.09} & \textbf{33.20} & 26.62 & \textbf{28.03} & \textbf{32.94} & \textbf{26.14} & \textbf{28.96}\\
    \midrule
     NeRF
    & 33.00 & 25.01 & 30.13 & 36.18
    & 32.54 & 29.62 & 32.91 & 28.65 & 31.01 \\
    \bottomrule
    \end{tabular*}
    \label{tab:psnr_nerf}
\end{table*}

Based on our radiance mapping, we further improve the performance by rectifying the raw coordinates of point clouds. One critical characteristics of point clouds is that it is generally noisy and sparse, due to unsatisfactory reconstruction or measurement quality. Specifically, the points $\mathbf{p}_{pcd}$ in a point cloud actually lie around the object surface $\mathbf{p}_{obj}$ with a spatial noise $\mathbf{\epsilon}$. We denote its coordinate with $\mathbf{x}_{pcd}=\mathbf{x}_{obj}+\mathbf{\epsilon}$. If we feed the raw point clouds into Eqn. \ref{equ:radiace_mapping}, we obtain $\mathbf{L}_{pcd} = \mathbf{F}_\Theta(\mathbf{x}_{obj}+\mathbf{\epsilon}, \mathbf{d})$, meaning that the queried latent code is actually from a nearby coordinate around the surface. The noise is mainly handled by adopting a coordinate rectification step before we feed them into the MLP. As shown in Figure \ref{fig:collapse} (b), after rasterization, we obtain the closest point to the camera position and its z-buffer for each pixel via depth test. We conduct ray marching to find the estimated intersection between the ray and the approximate surface. Specifically, we calculate the point which share the same z-buffer with the closest point and also lies on the ray:
\begin{equation}
    \mathbf{x}_{query} = \mathbf{o} + \frac{z_{pcd}}{cos\;\theta} \times \mathbf{d}
    \label{equ:coordinate_rectification}
\end{equation}
where $\mathbf{o}$ and $\mathbf{d}$ are the camera position and the normalized direction, and $\theta$ is the angle between the ray and the z axis pointing to the center of the scene. This simple approach has obvious benefits over prior art. This is because that using the raw point clouds would lead to spatial frequency collapse as conventional neural descriptors do. Namely, when the point clouds are sparse, the same point would be rasterized to multiple pixels. In other words, nearby pixels are highly probable to be queried to the same latent feature from the MLP mapper, leading to blurred results. Also, the point density of the raw point clouds has a non-uniform distribution on different parts. Thus the color of the closer points would have a larger impact on each other, as shown in Figure \ref{fig:collapse} (b). In contrast, our proposed rectification offers a more uniformly-distributed points density. Thus it results in more accurate geometry recovery. Empirical results also prove the effectiveness of this approach.

\subsection{The End-to-End Point Clouds Rendering Pipeline}


Our pipeline could be divided into three stages: rasterization, radiance mapping and refinement. In rasterization stage, we first transform the point clouds from world space to camera space. For each pixel, we search the points that lie within a radius threshold $\tau$ to the ray omitted from the camera to the pixel. The obtained points with their properties are so-called \textit{fragments}. Since this searching process is not required to be differentiable, it could be implemented by hardware-accelerated framework such as OpenGL (\citeauthor{shreiner2009opengl}) and ran in real time. In our experiments, we find the OpenGL-based implementation are 8-10 times faster than a pure software implementation, such as PyTorch3D (\citeauthor{johnson2020accelerating}). Based on the fragments obtained from the per-pixel search, we keep the closest point to the camera for the occupied pixels and abandon the rest, since we only evaluate the first intersected surface. Note that there are some unoccupied pixels that no points are found within the threshold $\tau$. These pixels leads to the artifact caused by the sparsity of point clouds and would be processed in the refinement stage. We transform the raw coordinates of the occupied pixels to the queried coordinates using the rectification in Eqn. \ref{equ:coordinate_rectification}. Then we transform the queried coordinates $\mathbf{x}_{query}=N_{occ} \times 3$ with position encoding introduced in NeRF (\citeauthor{mildenhall2020nerf}) and feed them into our spatial mapping function $\mathbf{F}_{\Theta}$ using Eqn. \ref{equ:radiace_mapping}, and obtain the latent feature $\mathbf{L}=N_{occ} \times C$. Then we re-organize these features back to a $H\times W\times C$ feature map, while we assign zero to the feature of those unoccupied pixels. The feature map output by MLP goes into the refinement stage. Unlike NeRF representation which uses a pure MLP design, the refinement stage is indispensable for point clouds rendering. The spatial mapping stage guarantees multi-view consistency, but outputs noisy 3D feature, while the refinement network is agnostic to the 3D geometry, but learn the texture of the images relying on the capability of convolution layers. The refinement stage is implemented as a 2D U-Net (\citeauthor{ronneberger2015u}). After the feature map is processed by the U-Net, we obtain the rendered image in shape of $H\times W\times 3$. We supervised the training by using the weighted sum of L2-loss and perceptual loss (\citeauthor{johnson2016perceptual, dosovitskiy2016generating}) as done in (\citeauthor{dai2020neural}), whose weights are $1$ and $0.01$ respectively.



\begin{figure*}[h!]
\begin{center}
  \includegraphics[width=\linewidth]{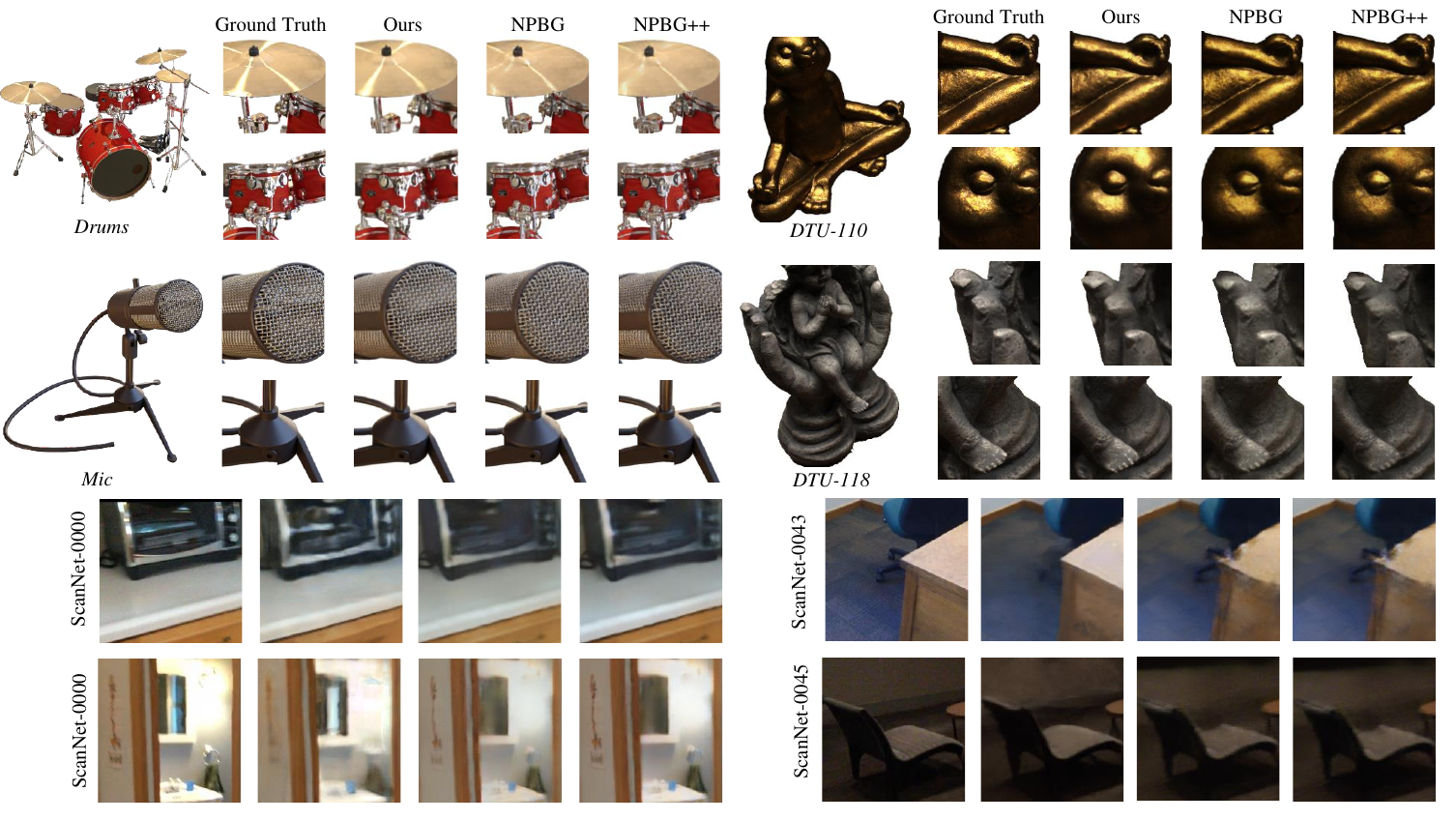}
\end{center}
\vspace{-12pt}
\caption{We compare the rendered novel views between our method and prior works. From the Drums scenes, we could see that the reflection rendered by our method is more photo-realistic than NPBG and NPBG++. Similar phenomenon is also observed in DTU-110. Besides, our method is able to produce elaborate details, such as the fine holes in the Mic scene, and the toes in DTU-118. We urge readers to zoom in the images for better observations.
}
\label{fig:result_vis}
\end{figure*}

\section{Experiments}


\subsection{Benchmark Evaluation}
\paragraph{Settings}
We experiment on three datasets: NeRF-Synthetic (\citeauthor{mildenhall2020nerf}), ScanNet (\citeauthor{dai2017scannet}) and DTU (\citeauthor{aanaes2016large}). We give a detailed description on them in the appendix. To speed up the training process, we only rasterize the point clouds once and save the fragments for reusing, since the training and testing views are fixed. We set the radius threshold $\tau$ as 5e-3 for NeRF-Synthetic, 1.5e-2 for ScanNet,
and 3e-3 for DTU. The selection of $\tau$ depends on the density of point clouds. If $\tau$ is set too large, the rasterized points would be inaccurate. If $\tau$ is set too small, then more pixels would become unoccupied, causing a so-called bleeding issue. We employ the same positional encoding to the input queried coordinates and ray directions as done in NeRF.  The output feature dimensions of each MLP layer are 256, 256, 256, 128 and 8 respectively. The input ray directions are concatenated with the output of second fully-connected layer. We use the same U-Net architecture as NPBG, where gated blocks are employed, consisting of gated convolution, ReLU and instance normalization layers. The U-Net down-samples the feature map for four times. The feature dimensions of each layer are 16, 32, 64, 128, 256. We use the Adam optimizer for training, with a batch size of 1. The initial learning rates of MLP and U-Net are 5e-4 and 1.5e-4 respectively, which are multiplied by 0.9999 in each step. We compare our methods with NPBG, NPBG++ and NeRF, using PSNR, SSIM and LPIPS as evaluation metrics. 

\paragraph{Results}
We show the quantitative results in Table \ref{tab:benchmark}. For NeRF-Synthetic, we report the results on Ficus, HotDog and Mic in Table \ref{tab:benchmark}, as done in NPBG++ (\citeauthor{rakhimov2022npbg++}). Besides, we additionally conduct experiments on all eight scenes and show their PSNR in Table \ref{tab:psnr_nerf}, while the rest metrics are shown in the appendix (Table \ref{tab:ssim_nerf}, \ref{tab:lpips_nerf}). Our method outperforms prior arts with notable margins on each dataset, in terms of PSNR, SSIM and LPIPS. Compared to NeRF, our method achieves comparable results on three datasets. It demonstrates the effectiveness of our method under various scenes. Table \ref{tab:psnr_nerf} show the PSNR on each scene in NeRF-Synthetic. Please refer to the appendix for all metrics. We could see our method achieves better results than NPBG and NPBG++ on 6 scenes out of 8, except for Lego and Ship. We visualize the results on these two scenes and their noisy point clouds in the appendix. In Figure \ref{fig:result_vis}, we illustrate the zoom-in views of four scenes. From the Drums scenes, we could see that the reflection rendered by our method is more photo-realistic than NPBG and NPBG++. Similar results are observed in DTU-110. Besides, our method is able to produce elaborate details, such as the fine holes in the Mic scene, and the toes in DTU-118. More results are presented in the appendix. We argue the reason that NPBG and NPBG++ lose fine details is the spatial frequency collapse and neighbor point disturbance issues caused by the discrete neural descriptors. Figure \ref{fig:drum} compare our method against NeRF. As analyzed before, our method and NeRF both model the translucency by optimizing the color of first-intersected surface. Our method avoids blurry artifact by leveraging the explicit geometry provided by point clouds. Besides, from the Materials scene, we observe that our method produces better specular reflection effect.

\begin{table*}[h!]
    \centering
    \setlength\tabcolsep{1pt}
    \caption{\textbf{(a)} We verify our method under various density of point clouds. ${1\times}$, ${10\times}$ and ${50\times}$ denote down-sampling the point clouds by 1, 10 and 50 times. \textit{w/o Rect.} denotes our method without coordinate rectification. The sparser the point cloud is, the more improvement brought by our method. \textbf{(b)} We analyze the influence of the refinement network capacity. \textit{Quad dim} denotes increasing the feature dimension by a factor of 4. Results are obtained using 400$\times$400 resolution due to GPU memory limitations. The three metrics are PSNR$\uparrow$, SSIM$\uparrow$ and LPIPS$\downarrow$ respectively.} 
    \vspace{-8pt}
    \begin{tabular*}{\hsize}{@{}@{\extracolsep{\fill}}lcccccccc@{}}
    \toprule
    Method & Chair $\uparrow/\uparrow/\downarrow$ & Drums $\uparrow/\uparrow/\downarrow$  & Lego $\uparrow/\uparrow/\downarrow$ & Mic $\uparrow/\uparrow/\downarrow$ \\
    \midrule
    
    NPBG++, ${1\times}$ & 29.33 / 0.950 / 0.045 & 24.68 / 0.919 / 0.067 & 27.27 / {0.905} / 0.080 & 32.72 / 0.971 / 0.026 \\
    
    Ours, ${1\times}$, w/o Rect. & 32.56 / 0.969 / 0.024 & 24.57 / 0.923 / 0.056 & 26.48 / 0.898 / 0.069 & 32.36 / 0.979 / 0.017 \\
    Ours, ${1\times}$ & 33.63 / 0.977 / 0.021 & 25.05 / 0.931 / 0.051 & {27.16} / 0.910 / 0.066 & 33.44 / 0.983 / 0.014 \\

    \midrule
    
    NPBG++, ${10\times}$ & 29.20 / 0.949 / 0.047 & 23.90 / 0.903 / 0.085 & 27.41 / 0.908 / 0.091 & 32.00 / 0.965 / 0.033 \\
    Ours, ${10\times}$, w/o Rect. & 31.24 / 0.958 / 0.030 & 23.77 / 0.910  / 0.069 & 25.88 / 0.887 / 0.083 & 31.42 / 0.975 / 0.022 \\
    Ours, ${10\times}$ & 32.46 / 0.967 / 0.028 & 24.26 / 0.919 / 0.065 & 26.46 / 0.900 / 0.077 & 32.91 / 0.981 / 0.016  \\
    
    \midrule
    
    NPBG++, ${50\times}$ & 27.82 / 0.929 / 0.058 & 22.72 / 0.876 / 0.104 & 25.72 / 0.871 / 0.119 & 31.21 / 0.959 / 0.039 \\
    Ours, ${50\times}$, w/o Rect. & 29.89 / 0.945 / 0.038 & 22.43 / 0.888 / 0.088 & 24.67 / 0.867 / 0.101 & 29.99 / 0.968 / 0.026 \\
    Ours, ${50\times}$ &  31.34 / 0.959 / 0.034 & 22.99 / 0.900 / 0.082 & 25.51 / 0.888 / 0.088 & 31.64 / 0.975 / 0.021 \\
    \midrule
    Ours, ${1\times}$, Qurd dim & 34.13 / 0.980 / 0.019 & 25.53 / 0.937 / 0.044 & 27.46 / 0.918 / 0.055 & 34.44 / 0.987 / 0.009 \\
    \bottomrule
    \end{tabular*}
    \label{tab:ablation}
\end{table*}

\begin{table}[!t]
    \centering
    \setlength\tabcolsep{1pt}
    \captionof{table}{Comparisons of model size and Frame Per Second (FPS). For point clouds rendering methods, the model size is seperated into spatial mapping and refinement stages. All results are measured on a TiTAN Xp.} 
    \vspace{-8pt}
    \begin{tabular*}{\hsize}{@{}@{\extracolsep{\fill}}lcc@{}}
    \toprule
    Method & Model Size $\downarrow$ &  FPS $\uparrow$\\
    \midrule
    NPBG  & (61+7.3)M &  17 \\
    NPBG++ & (17.5+1.2)M &  18 \\
    Ours & \textbf{(0.75+7.3)M} & \textbf{20} \\
    \midrule
    NeRF & \textbf{4.8M} &  0.015 \\
    Point-NeRF & 18.5M &  0.125 \\ 
    PlenOctrees & 440M & \textbf{110}  \\
    \bottomrule
    \end{tabular*}
    \label{tab:size_time}
\end{table}

\subsection{Complexity Analysis}
\textbf{(a) Model Size} Our model includes a 5-layer MLP for radiance mapping and a U-Net for refinement. Their sizes are 0.75M and 7.3M respectively. The NeRF MLP has a size of 4.8M. NPBG (\citeauthor{aliev2020neural}) uses the same U-Net architecture as ours (7.3M), but its neural descriptors are proportional to the number of points. Demonstrated with ScanNet, whose point clouds are about in the size of 2 millions points, the model size of the neural descriptors would be 61M. Hence the total size of NPBG is 68.3M. NPBG++ (\citeauthor{rakhimov2022npbg++}) uses a convolution neural network in the size of 17.5M as the image feature extractor to obtain the neural descriptor. Besides, NPBG++ uses a smaller U-Net in refinement stage, whose size is 1.2M. It results in a totol model size of 18.7M. PlenOctree has a much larger model size as 440M.
\textbf{(b) Time Complexity} In Table \ref{tab:size_time}, we show the Frame Per Second (FPS) of each method. All results are measured on a TiTAN Xp with an image size of $800\times 800$. The point clouds-based methods (ours, NPBG and NPBG++) require rasterization for pre-processing (All implemented in OpenGL). NPBG and NPBG++ need to rasterize the point clouds five times into five different resolution to feed into the refinement U-Net, while ours only need one. Therefore their FPS are slightly lower than ours. Our method outperforms these two methods with a FPS of 20. NeRF needs burdersome pre-processing, which severely slower the speed, yielding a 0.015 FPS. Point-NeRF follows the volume rendering formulation like NeRF with point clouds as input. Hence it also suffers from heavy sampling and high computation cost, with an FPS of 0.125.

\subsection{Ablation Study} 

\begin{figure}[t!]
\begin{center}
  \includegraphics[width=\linewidth]{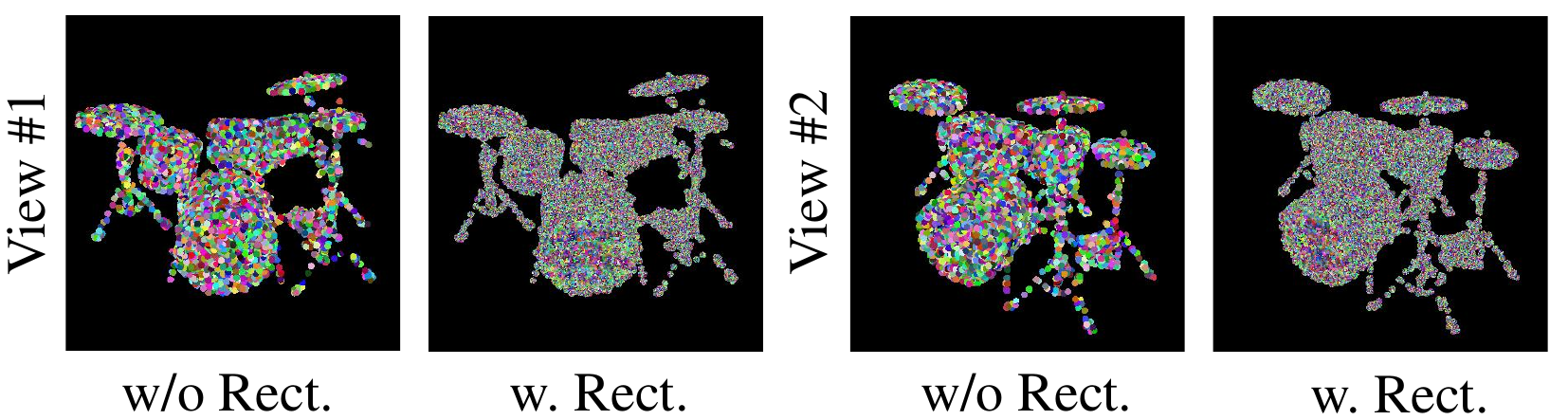}
\end{center}
\vspace{-8pt}
\caption{Rasterized feature of our method and baseline (w/o Rect.). Without coordinate rectification, the same neural feature would be rasterized into multiple pixels and optimized to fit the average of these pixels, which would harms the rendering quality. Our method could notably allivete the issue.}
\label{fig:rasterized_feature}
\end{figure}

\paragraph{Point clouds Density}  We evaluate our methods under different point clouds density. We select four scenes from NeRF-Synthetic dataset (Chair, Drums, Lego and Mic) for evaluation. We conduct these experiments with a $400\times400$ resolution, without affecting the conclusion. Specifically, we uniformly down-sample the original point clouds by a factor of 10 and 50 respectively, to explore how much improvement the proposed method brings under different point cloud density. We compare our method against  NPBG++ and our baseline without coordinate rectification. We show the PSNR, SSIM, and LPIPS in Table \ref{tab:ablation}. It's seen that our method achieves best performance in almost all cases. Moreover, the sparser the point cloud is, the greater improvement is brought by our method over NPBG++ and the baseline. The reason is that in NPBG++ and the baseline, the same neural feature would be rasterized into multiple pixels and optimized to fit the average of these pixels, which would harms the rendering quality. When the point cloud becomes sparser, this issue becomes more serious. Our method with coordinate rectification could notably alleviate this issue. In Figure \ref{fig:rasterized_feature}, we illustrate the rasterized feature of our method and the baseline, regarding three channels as RGB. NPBG and NPBG++ has similar effect as baseline (w/o Rect.) does.

\paragraph{Capacity of Refinement Network} Since the refinement network is actually playing the role of denoising or inpainting, it would be important to know how the capacity of the network influence the image quality. We increase the channel dimension in each layer of the refinement network by a factor of four, yielding a model size of 127M. As shown in the bottom row in Table \ref{tab:ablation}, increasing the capacity further improve our performance.



\section{Conclusion}

In this work, we propose a point clouds rendering method that achieve photo-realistic rendering quality. We begin with revisiting the spatial mapping functions in prior art and then propose our radiance mapping based on the adaptation of volumetric rendering formulation on point clouds. Further, we propose coordinate rectification which alleviate the spatial frequency collapse and neighbor point disturbance issues. Composed of rasterization, spatial mapping and the refinement stages, Our method achieves the state-of-the-art performance on point clouds rendering task.

\section{Acknowledgement}

This work was supported by National Science Foundation of China (U20B2072, 61976137). This work was also partially supported by Grant YG2021ZD18 from Shanghai Jiao Tong University Medical Engineering Cross Research. This work was also partially supported by the Shanghai Committee of Science and Technology, China (Grant No. 20DZ1100800)

\bibliography{aaai23}

\begin{thebibliography}{43}
\providecommand{\natexlab}[1]{#1}

\bibitem[{Aan{\ae}s et~al.(2016)Aan{\ae}s, Jensen, Vogiatzis, Tola, and
  Dahl}]{aanaes2016large}
Aan{\ae}s, H.; Jensen, R.~R.; Vogiatzis, G.; Tola, E.; and Dahl, A.~B. 2016.
\newblock Large-scale data for multiple-view stereopsis.
\newblock \emph{International Journal of Computer Vision}, 120(2): 153--168.

\bibitem[{Aliev et~al.(2020)Aliev, Sevastopolsky, Kolos, Ulyanov, and
  Lempitsky}]{aliev2020neural}
Aliev, K.-A.; Sevastopolsky, A.; Kolos, M.; Ulyanov, D.; and Lempitsky, V.
  2020.
\newblock Neural point-based graphics.
\newblock In \emph{European Conference on Computer Vision}, 696--712. Springer.

\bibitem[{Barron et~al.(2021)Barron, Mildenhall, Tancik, Hedman,
  Martin-Brualla, and Srinivasan}]{barron2021mip}
Barron, J.~T.; Mildenhall, B.; Tancik, M.; Hedman, P.; Martin-Brualla, R.; and
  Srinivasan, P.~P. 2021.
\newblock Mip-nerf: A multiscale representation for anti-aliasing neural
  radiance fields.
\newblock In \emph{Proceedings of the IEEE/CVF International Conference on
  Computer Vision}, 5855--5864.

\bibitem[{Chen et~al.(2022)Chen, Xu, Geiger, Yu, and Su}]{chen2022tensorf}
Chen, A.; Xu, Z.; Geiger, A.; Yu, J.; and Su, H. 2022.
\newblock TensoRF: Tensorial Radiance Fields.
\newblock \emph{arXiv preprint arXiv:2203.09517}.

\bibitem[{Dai et~al.(2017)Dai, Chang, Savva, Halber, Funkhouser, and
  Nie{\ss}ner}]{dai2017scannet}
Dai, A.; Chang, A.~X.; Savva, M.; Halber, M.; Funkhouser, T.; and Nie{\ss}ner,
  M. 2017.
\newblock Scannet: Richly-annotated 3d reconstructions of indoor scenes.
\newblock In \emph{Proceedings of the IEEE conference on computer vision and
  pattern recognition}, 5828--5839.

\bibitem[{Dai et~al.(2020)Dai, Zhang, Li, Liu, and Zeng}]{dai2020neural}
Dai, P.; Zhang, Y.; Li, Z.; Liu, S.; and Zeng, B. 2020.
\newblock Neural point cloud rendering via multi-plane projection.
\newblock In \emph{Proceedings of the IEEE/CVF Conference on Computer Vision
  and Pattern Recognition}, 7830--7839.

\bibitem[{Dosovitskiy and Brox(2016)}]{dosovitskiy2016generating}
Dosovitskiy, A.; and Brox, T. 2016.
\newblock Generating images with perceptual similarity metrics based on deep
  networks.
\newblock \emph{Advances in neural information processing systems}, 29.

\bibitem[{Godard, Mac~Aodha, and Brostow(2017)}]{godard2017unsupervised}
Godard, C.; Mac~Aodha, O.; and Brostow, G.~J. 2017.
\newblock Unsupervised monocular depth estimation with left-right consistency.
\newblock In \emph{Proceedings of the IEEE conference on computer vision and
  pattern recognition}, 270--279.

\bibitem[{He et~al.(2016)He, Zhang, Ren, and Sun}]{he2016deep}
He, K.; Zhang, X.; Ren, S.; and Sun, J. 2016.
\newblock Deep residual learning for image recognition.
\newblock In \emph{Proceedings of the IEEE conference on computer vision and
  pattern recognition}, 770--778.

\bibitem[{Huang et~al.(2022)Huang, Yang, Wang, Chen, Li, Li, Ni, and
  Zhang}]{huang2022representation}
Huang, X.; Yang, J.; Wang, Y.; Chen, Z.; Li, L.; Li, T.; Ni, B.; and Zhang, W.
  2022.
\newblock Representation-Agnostic Shape Fields.
\newblock In \emph{International Conference on Learning Representations}.

\bibitem[{Insafutdinov and Dosovitskiy(2018)}]{insafutdinov2018unsupervised}
Insafutdinov, E.; and Dosovitskiy, A. 2018.
\newblock Unsupervised learning of shape and pose with differentiable point
  clouds.
\newblock \emph{Advances in neural information processing systems}, 31.

\bibitem[{Jain, Tancik, and Abbeel(2021)}]{jain2021putting}
Jain, A.; Tancik, M.; and Abbeel, P. 2021.
\newblock Putting nerf on a diet: Semantically consistent few-shot view
  synthesis.
\newblock In \emph{Proceedings of the IEEE/CVF International Conference on
  Computer Vision}, 5885--5894.

\bibitem[{Johnson, Alahi, and Fei-Fei(2016)}]{johnson2016perceptual}
Johnson, J.; Alahi, A.; and Fei-Fei, L. 2016.
\newblock Perceptual losses for real-time style transfer and super-resolution.
\newblock In \emph{European conference on computer vision}, 694--711. Springer.

\bibitem[{Johnson et~al.(2020)Johnson, Ravi, Reizenstein, Novotny, Tulsiani,
  Lassner, and Branson}]{johnson2020accelerating}
Johnson, J.; Ravi, N.; Reizenstein, J.; Novotny, D.; Tulsiani, S.; Lassner, C.;
  and Branson, S. 2020.
\newblock Accelerating 3d deep learning with pytorch3d.
\newblock In \emph{SIGGRAPH Asia 2020 Courses}, 1--1.

\bibitem[{Kopanas et~al.(2021)Kopanas, Philip, Leimk{\"u}hler, and
  Drettakis}]{kopanas2021point}
Kopanas, G.; Philip, J.; Leimk{\"u}hler, T.; and Drettakis, G. 2021.
\newblock Point-Based Neural Rendering with Per-View Optimization.
\newblock In \emph{Computer Graphics Forum}, volume~40, 29--43. Wiley Online
  Library.

\bibitem[{Li et~al.(2022)Li, Li, Ma, Zhang, Chen, and Zhu}]{li2022read}
Li, Z.; Li, L.; Ma, Z.; Zhang, P.; Chen, J.; and Zhu, J. 2022.
\newblock READ: Large-Scale Neural Scene Rendering for Autonomous Driving.
\newblock arXiv:2205.05509.

\bibitem[{Lin, Kong, and Lucey(2018)}]{lin2018learning}
Lin, C.-H.; Kong, C.; and Lucey, S. 2018.
\newblock Learning efficient point cloud generation for dense 3d object
  reconstruction.
\newblock In \emph{proceedings of the AAAI Conference on Artificial
  Intelligence}, volume~32.

\bibitem[{Liu et~al.(2020)Liu, Gu, Zaw~Lin, Chua, and Theobalt}]{liu2020neural}
Liu, L.; Gu, J.; Zaw~Lin, K.; Chua, T.-S.; and Theobalt, C. 2020.
\newblock Neural sparse voxel fields.
\newblock \emph{Advances in Neural Information Processing Systems}, 33:
  15651--15663.

\bibitem[{Martin-Brualla et~al.(2021)Martin-Brualla, Radwan, Sajjadi, Barron,
  Dosovitskiy, and Duckworth}]{martin2021nerf}
Martin-Brualla, R.; Radwan, N.; Sajjadi, M.~S.; Barron, J.~T.; Dosovitskiy, A.;
  and Duckworth, D. 2021.
\newblock Nerf in the wild: Neural radiance fields for unconstrained photo
  collections.
\newblock In \emph{Proceedings of the IEEE/CVF Conference on Computer Vision
  and Pattern Recognition}, 7210--7219.

\bibitem[{Maturana and Scherer(2015)}]{maturana2015voxnet}
Maturana, D.; and Scherer, S. 2015.
\newblock Voxnet: A 3d convolutional neural network for real-time object
  recognition.
\newblock In \emph{2015 IEEE/RSJ international conference on intelligent robots
  and systems (IROS)}, 922--928. IEEE.

\bibitem[{Mildenhall et~al.(2020)Mildenhall, Srinivasan, Tancik, Barron,
  Ramamoorthi, and Ng}]{mildenhall2020nerf}
Mildenhall, B.; Srinivasan, P.~P.; Tancik, M.; Barron, J.~T.; Ramamoorthi, R.;
  and Ng, R. 2020.
\newblock Nerf: Representing scenes as neural radiance fields for view
  synthesis.
\newblock In \emph{European conference on computer vision}, 405--421. Springer.

\bibitem[{Niemeyer et~al.(2020)Niemeyer, Mescheder, Oechsle, and
  Geiger}]{niemeyer2020differentiable}
Niemeyer, M.; Mescheder, L.; Oechsle, M.; and Geiger, A. 2020.
\newblock Differentiable volumetric rendering: Learning implicit 3d
  representations without 3d supervision.
\newblock In \emph{Proceedings of the IEEE/CVF Conference on Computer Vision
  and Pattern Recognition}, 3504--3515.

\bibitem[{Oechsle, Peng, and Geiger(2021)}]{oechsle2021unisurf}
Oechsle, M.; Peng, S.; and Geiger, A. 2021.
\newblock Unisurf: Unifying neural implicit surfaces and radiance fields for
  multi-view reconstruction.
\newblock In \emph{Proceedings of the IEEE/CVF International Conference on
  Computer Vision}, 5589--5599.

\bibitem[{Ost et~al.(2021)Ost, Laradji, Newell, Bahat, and
  Heide}]{DBLP:journals/corr/abs-2112-01473}
Ost, J.; Laradji, I.; Newell, A.; Bahat, Y.; and Heide, F. 2021.
\newblock Neural Point Light Fields.
\newblock \emph{CoRR}, abs/2112.01473.

\bibitem[{Rakhimov et~al.(2022)Rakhimov, Ardelean, Lempitsky, and
  Burnaev}]{rakhimov2022npbg++}
Rakhimov, R.; Ardelean, A.-T.; Lempitsky, V.; and Burnaev, E. 2022.
\newblock NPBG++: Accelerating Neural Point-Based Graphics.
\newblock In \emph{Proceedings of the IEEE conference on computer vision and
  pattern recognition}.

\bibitem[{Ronneberger, Fischer, and Brox(2015)}]{ronneberger2015u}
Ronneberger, O.; Fischer, P.; and Brox, T. 2015.
\newblock U-net: Convolutional networks for biomedical image segmentation.
\newblock In \emph{International Conference on Medical image computing and
  computer-assisted intervention}, 234--241. Springer.

\bibitem[{R{\"u}ckert, Franke, and Stamminger(2021)}]{ruckert2021adop}
R{\"u}ckert, D.; Franke, L.; and Stamminger, M. 2021.
\newblock Adop: Approximate differentiable one-pixel point rendering.
\newblock \emph{arXiv preprint arXiv:2110.06635}.

\bibitem[{Shreiner, Group et~al.(2009)}]{shreiner2009opengl}
Shreiner, D.; Group, B. T. K. O. A.~W.; et~al. 2009.
\newblock \emph{OpenGL programming guide: the official guide to learning
  OpenGL, versions 3.0 and 3.1}.
\newblock Pearson Education.

\bibitem[{Wang et~al.(2021{\natexlab{a}})Wang, Galliani, Vogel, Speciale, and
  Pollefeys}]{wang2021patchmatchnet}
Wang, F.; Galliani, S.; Vogel, C.; Speciale, P.; and Pollefeys, M.
  2021{\natexlab{a}}.
\newblock Patchmatchnet: Learned multi-view patchmatch stereo.
\newblock In \emph{Proceedings of the IEEE/CVF Conference on Computer Vision
  and Pattern Recognition}, 14194--14203.

\bibitem[{Wang et~al.(2021{\natexlab{b}})Wang, Liu, Liu, Theobalt, Komura, and
  Wang}]{wang2021neus}
Wang, P.; Liu, L.; Liu, Y.; Theobalt, C.; Komura, T.; and Wang, W.
  2021{\natexlab{b}}.
\newblock NeuS: Learning Neural Implicit Surfaces by Volume Rendering for
  Multi-view Reconstruction.
\newblock In \emph{Advances in Neural Information Processing Systems}.

\bibitem[{Wang et~al.(2021{\natexlab{c}})Wang, Wang, Genova, Srinivasan, Zhou,
  Barron, Martin-Brualla, Snavely, and Funkhouser}]{wang2021ibrnet}
Wang, Q.; Wang, Z.; Genova, K.; Srinivasan, P.~P.; Zhou, H.; Barron, J.~T.;
  Martin-Brualla, R.; Snavely, N.; and Funkhouser, T. 2021{\natexlab{c}}.
\newblock Ibrnet: Learning multi-view image-based rendering.
\newblock In \emph{Proceedings of the IEEE/CVF Conference on Computer Vision
  and Pattern Recognition}, 4690--4699.

\bibitem[{Wang et~al.(2018)Wang, Liu, Zhu, Tao, Kautz, and
  Catanzaro}]{wang2018high}
Wang, T.-C.; Liu, M.-Y.; Zhu, J.-Y.; Tao, A.; Kautz, J.; and Catanzaro, B.
  2018.
\newblock High-resolution image synthesis and semantic manipulation with
  conditional gans.
\newblock In \emph{Proceedings of the IEEE conference on computer vision and
  pattern recognition}, 8798--8807.

\bibitem[{Wiles et~al.(2020)Wiles, Gkioxari, Szeliski, and
  Johnson}]{wiles2020synsin}
Wiles, O.; Gkioxari, G.; Szeliski, R.; and Johnson, J. 2020.
\newblock Synsin: End-to-end view synthesis from a single image.
\newblock In \emph{Proceedings of the IEEE/CVF Conference on Computer Vision
  and Pattern Recognition}, 7467--7477.

\bibitem[{Xu et~al.(2022)Xu, Xu, Philip, Bi, Shu, Sunkavalli, and
  Neumann}]{xu2022point}
Xu, Q.; Xu, Z.; Philip, J.; Bi, S.; Shu, Z.; Sunkavalli, K.; and Neumann, U.
  2022.
\newblock Point-NeRF: Point-based Neural Radiance Fields.
\newblock \emph{arXiv preprint arXiv:2201.08845}.

\bibitem[{Yang et~al.(2021)Yang, Huang, He, Xu, Yang, Xu, and
  Ni}]{yang2021reinventing}
Yang, J.; Huang, X.; He, Y.; Xu, J.; Yang, C.; Xu, G.; and Ni, B. 2021.
\newblock Reinventing 2d convolutions for 3d images.
\newblock \emph{IEEE Journal of Biomedical and Health Informatics}, 25(8):
  3009--3018.

\bibitem[{Yao et~al.(2018)Yao, Luo, Li, Fang, and Quan}]{yao2018mvsnet}
Yao, Y.; Luo, Z.; Li, S.; Fang, T.; and Quan, L. 2018.
\newblock Mvsnet: Depth inference for unstructured multi-view stereo.
\newblock In \emph{Proceedings of the European Conference on Computer Vision
  (ECCV)}, 767--783.

\bibitem[{Yariv et~al.(2021)Yariv, Gu, Kasten, and Lipman}]{yariv2021volume}
Yariv, L.; Gu, J.; Kasten, Y.; and Lipman, Y. 2021.
\newblock Volume rendering of neural implicit surfaces.
\newblock \emph{Advances in Neural Information Processing Systems}, 34.

\bibitem[{Yariv et~al.(2020)Yariv, Kasten, Moran, Galun, Atzmon, Ronen, and
  Lipman}]{yariv2020multiview}
Yariv, L.; Kasten, Y.; Moran, D.; Galun, M.; Atzmon, M.; Ronen, B.; and Lipman,
  Y. 2020.
\newblock Multiview neural surface reconstruction by disentangling geometry and
  appearance.
\newblock \emph{Advances in Neural Information Processing Systems}, 33:
  2492--2502.

\bibitem[{Yifan et~al.(2019)Yifan, Serena, Wu, {\"O}ztireli, and
  Sorkine-Hornung}]{yifan2019differentiable}
Yifan, W.; Serena, F.; Wu, S.; {\"O}ztireli, C.; and Sorkine-Hornung, O. 2019.
\newblock Differentiable surface splatting for point-based geometry processing.
\newblock \emph{ACM Transactions on Graphics (TOG)}, 38(6): 1--14.

\bibitem[{Yu et~al.(2021{\natexlab{a}})Yu, Li, Tancik, Li, Ng, and
  Kanazawa}]{yu2021plenoctrees}
Yu, A.; Li, R.; Tancik, M.; Li, H.; Ng, R.; and Kanazawa, A.
  2021{\natexlab{a}}.
\newblock Plenoctrees for real-time rendering of neural radiance fields.
\newblock In \emph{Proceedings of the IEEE/CVF International Conference on
  Computer Vision}, 5752--5761.

\bibitem[{Yu et~al.(2021{\natexlab{b}})Yu, Ye, Tancik, and
  Kanazawa}]{yu2021pixelnerf}
Yu, A.; Ye, V.; Tancik, M.; and Kanazawa, A. 2021{\natexlab{b}}.
\newblock pixelnerf: Neural radiance fields from one or few images.
\newblock In \emph{Proceedings of the IEEE/CVF Conference on Computer Vision
  and Pattern Recognition}, 4578--4587.

\bibitem[{Zhou et~al.(2017)Zhou, Brown, Snavely, and
  Lowe}]{zhou2017unsupervised}
Zhou, T.; Brown, M.; Snavely, N.; and Lowe, D.~G. 2017.
\newblock Unsupervised learning of depth and ego-motion from video.
\newblock In \emph{Proceedings of the IEEE conference on computer vision and
  pattern recognition}, 1851--1858.

\bibitem[{Zwicker et~al.(2001)Zwicker, Pfister, Van~Baar, and
  Gross}]{zwicker2001surface}
Zwicker, M.; Pfister, H.; Van~Baar, J.; and Gross, M. 2001.
\newblock Surface splatting.
\newblock In \emph{Proceedings of the 28th annual conference on Computer
  graphics and interactive techniques}, 371--378.

\end{thebibliography}
\clearpage
\appendix
\begin{appendices}
\section{Appendix A: Datasets}

\paragraph{NeRF-Synthetic} 
NeRF-Synthetic (\citeauthor{mildenhall2020nerf}) is a high quality synthetic dataset containing pathtraced images of 8 objects that exhibit complex geometry and realistic non-Lambertian materials. 
For each object, there are 100 images for training and 200 images for testing.
We use the point clouds provided by Point-NeRF (\citeauthor{xu2022point}). During training stage, we conduct data augmentation by random scaling and cropping as done in prior works. We set the training window size as 800$\times$800.
At test time, we render novel view images with the original resolution 800$\times$800.
\paragraph{ScanNet} ScanNet (\citeauthor{dai2017scannet}) is a RGBD dataset of indoor scenes. We evaluate our method on three scenes \{\textit{scene0000\_00}, \textit{scene0043\_00} and \textit{scene0045\_00}\} as done in NPBG++ (\citeauthor{rakhimov2022npbg++}) and follow their train-test-split. Specifically, if there are less than 2000 frames in the scene, we take 100 frames with equal spacing in the image stream. If there are more than 2000 frames, we take every 20th image such that the training views would not be too sparse. \textit{scene0000\_00} belongs to the 
latter case and the other two belong to the former. Then, we sample 10 frames at a fixed interval for testing and the rest for training. The training window size is set to 720$\times$720 and the test resolution is 960$\times$1200.

\paragraph{DTU} 
DTU (\citeauthor{aanaes2016large}) is a multi-view stereo dataset captured at a resolution of 1200 $\times$ 1600. We validate our approach on three scenes \{\textit{scan110}, \textit{scan114}, \textit{scan118}\} as done in NPBG++ (\citeauthor{rakhimov2022npbg++}) and use the same point clouds as theirs, which are obtained by PatchMatchNet (\citeauthor{wang2021patchmatchnet}). We also perform the same train-test-split in each scene. NPBG++ evaluate the whole images including the background. In our empirical observations, we find that the background is dark and far from the camera, producing a extremely noisy point cloud. Although the models are able to render high-quality details of the objects, the numerical results are still very low. We argue that such evaluation settings do not match our visual effects and is not reasonable. Therefore, we mask out the background both in training and testing using the binary segmentation masks provided by IDR (\citeauthor{yariv2020multiview}). All methods are evaluated using this settings. The training window size is set to 800$\times$800 and the test resolution 1200 $\times$ 1600.

\section{Appendix B: Limitation}
The proposed method produces photo-realistic images based on point clouds, whose performance is comparable to NeRF. However, similar to previous point clouds rendering methods, our performance is restricted by the point clouds quality. Specifically, when the point clouds quality is extremely poor, such as the one in the Lego scene in NeRF-Synthetic, our performance would degrade severely, while NeRF is hardly affected. Besides, although our method produces better visual effects in some cases such as specular reflection compared to NeRF, we does not outperform NeRF in terms of numerical results.


\begin{figure*}[ht!]
\begin{center}
  \includegraphics[width=\linewidth]{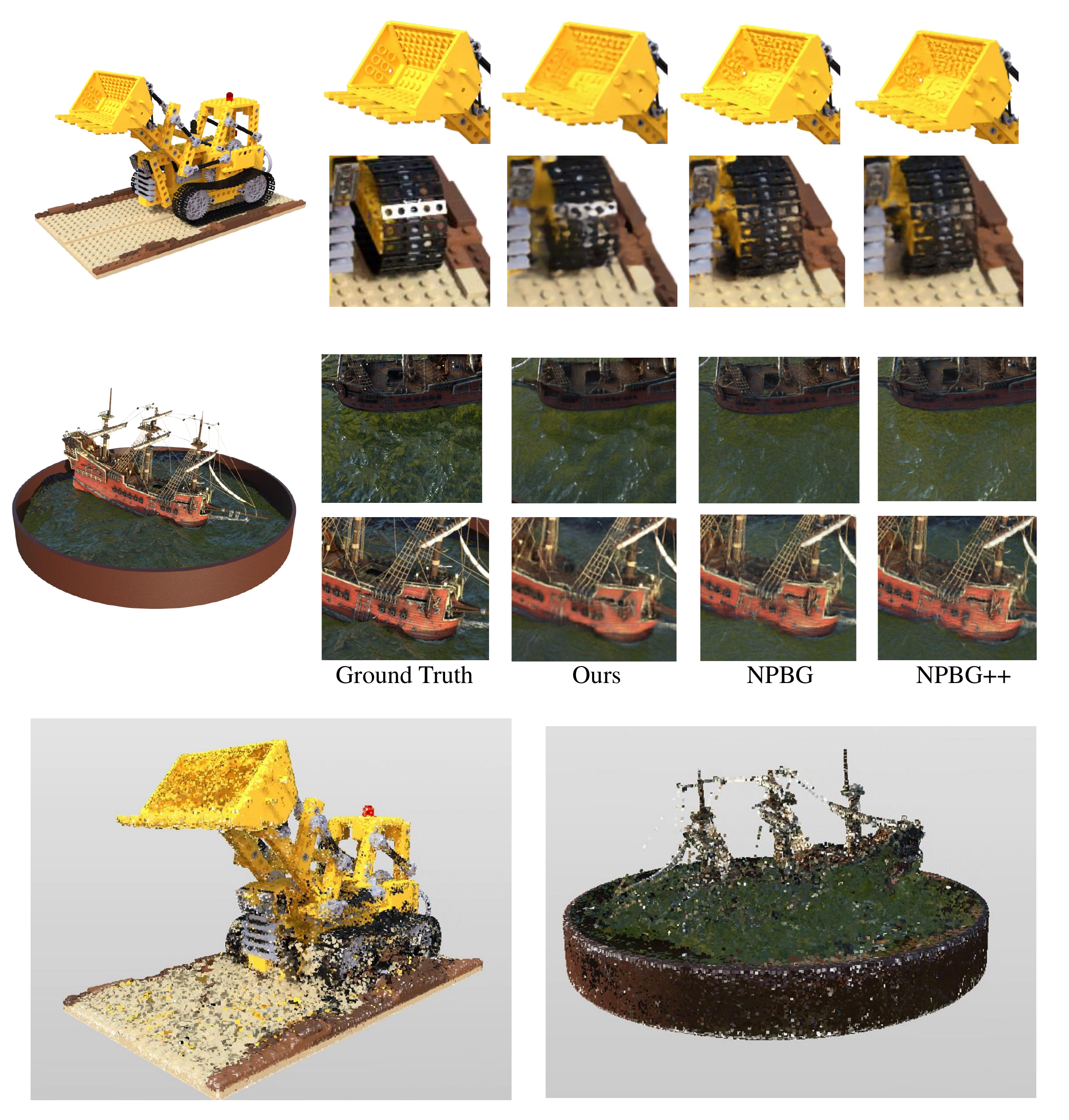}
\end{center}
\caption{Qualitative results on Lego and Ship scenes. In Lego scene, a layer of points are wrongly reconstructed by MVSNet, which lie ahead of the real surface on the shovel (See first row images). With the wrong geometry, fitting one view well might result in 3D inconsistency in another view, leading to unusual artifact for all three point-based methods. The second row images of Lego shows that our method produce better specular effect compared to NPBG and NPBG++. We also show the noisy point clouds of these two scenes for demonstration.}
\label{fig:lego_ship}
\end{figure*}



\begin{table*}[!h]
    \centering
    \setlength\tabcolsep{1pt}
    \caption{SSIM$\uparrow$ on each scene in NeRF-Synthetic}
    \begin{tabular*}{\hsize}{@{}@{\extracolsep{\fill}}lccccccccc@{}}
    \toprule
    Method & Chair & Drums & Ficus & Hotdog & Lego & Materials & Mic & Ship & Mean\\
    \midrule
    
    NPBG 
    & 0.954 & 0.902 & 0.942 & 0.960 & 0.919 & 0.922 & 0.970 & 0.812 & 0.923   \\
    NPBG++ 
    & \textbf{0.961} & 0.910 & 0.947 & 0.961 & \textbf{0.923} & 0.925 & 0.972 & 0.822 & 0.928  \\
    Ours 
    & 0.953 & \textbf{0.924} & \textbf{0.958} & \textbf{0.964} & 0.902 & \textbf{0.945} & \textbf{0.983} & \textbf{0.824} & \textbf{0.932}  \\
    \midrule
    NeRF 
    & 0.992 & 0.956 & 0.985 & 0.991
    & 0.988 & 0.982 & 0.992 & 0.928 & 0.977\\
    \bottomrule
    \end{tabular*}
    \label{tab:ssim_nerf}
\end{table*}
    
\begin{table*}[!h]
    \centering
    \setlength\tabcolsep{1pt}
    \caption{LPIPS$\downarrow$ on each scene in NeRF-Synthetic}
    \begin{tabular*}{\hsize}{@{}@{\extracolsep{\fill}}lccccccccc@{}}
    \toprule
    Method & Chair & Drums & Ficus & Hotdog & Lego & Materials & Mic & Ship & Mean\\
    \midrule
    
    NPBG 
    & 0.047 & 0.093 & 0.046 & 0.055 & 0.089 & 0.076 & 0.037 & 0.171 & 0.077    \\
    NPBG++ 
    & 0.048 & 0.092 & 0.043 & 0.053 & 0.089 & 0.074 & 0.030 & 0.178 & 0.076 \\
    Ours 
    & \textbf{0.040} & \textbf{0.068} & \textbf{0.035} & \textbf{0.038} & \textbf{0.085} & \textbf{0.050} & \textbf{0.014} & \textbf{0.159} & \textbf{0.061} \\
    \midrule
    NeRF 
    & 0.046 & 0.091 & 0.044 & 0.121 
    & 0.050 & 0.063 & 0.028 & 0.206 & 0.081\\
    \bottomrule
    \end{tabular*}
    \label{tab:lpips_nerf}
\end{table*}

\begin{table*}[!h]
    \centering
    \setlength\tabcolsep{1pt}
    \caption{PSNR$\uparrow$ on each scene in ScanNet and DTU}
    \begin{tabular*}{\hsize}{@{}@{\extracolsep{\fill}}lcccccc@{}}
    \toprule
    Method & ScanNet-0000 & ScanNet-0043 & ScanNet-0045 & DTU-110 & DTU-114 & DTU-118 \\
    \midrule
    
    NPBG
    & 22.24 & 25.27 & 27.75 & 29.20 & \textbf{28.05} & 33.47\\
    NPBG++ 
    & 22.25 & \textbf{25.43} & 28.03 & 29.50 & 27.84 & 33.43\\
    Ours 
    & \textbf{23.79} & 25.26 & \textbf{28.59} & \textbf{30.79} & 27.91 & \textbf{33.72}\\
    \midrule
    NeRF
    & 22.10 & 25.64 & 29.16 & 30.32 & 28.38 & 34.38\\
    \bottomrule
    \end{tabular*}
    \label{tab:psnr_scandtu}
\end{table*}

\begin{table*}[!h]
    \centering
    \setlength\tabcolsep{1pt}
    \caption{SSIM$\uparrow$ on each scene in ScanNet and DTU}
    \begin{tabular*}{\hsize}{@{}@{\extracolsep{\fill}}lcccccc@{}}
    \toprule
    Method & ScanNet-0000 & ScanNet-0043 & ScanNet-0045 & DTU-110 & DTU-114 & DTU-118 \\
    \midrule
    
    NPBG
    & 0.695 & 0.830 & 0.686 & 0.907 & \textbf{0.873} & 0.904  \\
    NPBG++ 
    & 0.712 & 0.821 & 0.731 & 0.910 & 0.865 & 0.909 \\
    Ours 
    & \textbf{0.748} & \textbf{0.844} & \textbf{0.789} & \textbf{0.924} & 0.863 & \textbf{0.926} \\
    \midrule
    NeRF 
    & 0.710 & 0.833 & 0.753 & 0.916 & 0.870 & 0.942 \\
    \bottomrule
    \end{tabular*}
    \label{tab:ssim_scandtu}
\end{table*}

\begin{table*}[!h]
    \centering
    \setlength\tabcolsep{1pt}
    \caption{LPIPS$\downarrow$ on each scene in ScanNet and DTU}
    \begin{tabular*}{\hsize}{@{}@{\extracolsep{\fill}}lcccccc@{}}
    \toprule
    Method & ScanNet-0000 & ScanNet-0043 & ScanNet-0045 & DTU-110 & DTU-114 & DTU-118 \\
    \midrule
    
    NPBG 
    & 0.474 & 0.421 & 0.482 & 0.124 & \textbf{0.143} & 0.123\\
    NPBG++ 
    & 0.456  & 0.412 & 0.465 & 0.125 & 0.148 & 0.121\\
    Ours 
    & \textbf{0.440} & \textbf{0.389} & \textbf{0.415} & \textbf{0.114} & 0.154 & \textbf{0.117} \\
    \midrule
    NeRF 
    & 0.461 & 0.402 & 0.447 & 0.124 & 0.140 & 0.116   \\
    \bottomrule
    \end{tabular*}
    \label{tab:lpips_scandtu}
\end{table*}
\end{appendices}




\end{document}